# Multimodal fusion of visual and morphometric features for avian bone classification

Nevio Dubbini[1], Lisa Yeomans[2], Marco Pavia[3], Ramazan Parmaksiz[3], Ayse Atas Hooglugt[4], Gabriele Gattiglia[1], Beatrice Demarchi[3]

[1] University of Pisa, Italy, nevio.dubbini@unipi.it
[2] University of Copenhagen, Denmark
[3] University of Turin, Italy
[4] University of Groningen, Groningen, Netherlands

**Abstract**. Artificial intelligence has shown considerable potential for archaeological applications, yet its use in zooarchaeology remains limited, particularly for the identification of avian skeletal remains. This study presents a proof-of-concept multimodal framework that integrates convolutional neural network-based image analysis with osteometric measurements for the classification of bird bones. Using a dataset of more than 10,000 images from multiple museum and research collections, two classification tasks were investigated: skeletal element identification and family-level taxonomic classification. Prior to classification, images were automatically segmented using a two-stage pipeline combining BiRefNet and SAM2. Visual features extracted with a pre-trained EfficientNet_V2_S backbone were fused with standardized morphometric data through a feature-level multimodal architecture. The model achieved 86% accuracy on the test set for bone-type classification, demonstrating reliable recognition of skeletal elements. Family-level classification proved more challenging, reaching 51% top-1 accuracy but 75% top-3 accuracy, indicating that correct taxa were frequently included among the most probable predictions. These results demonstrate the feasibility of combining visual and morphometric information within a unified deep-learning framework and establish a methodological baseline for future AI-assisted zooarchaeological identification. The approach contributes to ongoing efforts to develop scalable, interpretable, and archaeologically meaningful tools for the study of avian remains.

# Introduction

Artificial Intelligence (AI) is increasingly integrated into archaeology, where it is used to relieve human operators from repetitive and time-consuming tasks, process large digital datasets, and operate in difficult or inaccessible environments (Argyrou & Agapiou, 2022; Chapinal-Heras & Díaz-Sánchez, 2023). AI has become widely adopted for image-related applications such as classification, detection, and segmentation (Goodfellow et al., 2016). In archaeology, AI has shown great potential for the study of landscapes, sites, artefacts and even ecofacts (Orengo et al., 2020; Caspari & Crespo, 2019; Verschoof-van der Vaart & Lambers, 2022; Küçükdemirci & Sarris, 2020; Anichini et al., 2021; Gualandi et al., 2021; Resler et al., 2021; Tsigkas et al., 2020; Pierdica et al., 2020; Tanti et al., 2021), such as historical cartography archaeobotanical remains, human remains, and bone surface modifications (Bewes et al., 2019; Bonhomme et al., 2023; Dominguez-Rodrigo et al., 2020).

The rapid growth of data generated across heterogeneous modalities has driven increasing interest in multimodal artificial intelligence. Unlike traditional machine learning approaches that process a single data modality in isolation, multimodal AI seeks to jointly exploit complementary information from different sources, enabling a more comprehensive understanding of complex phenomena and improving predictive performance (Barua et al., 2023; Liang et al., 2024). Over the last decade, a substantial body of literature has demonstrated the effectiveness of multimodal approaches in a wide range of domains, including natural language processing, healthcare, autonomous driving, recommendation systems in e-commerce and predictive maintenance and quality control in industrial environments (Gandhu et al., 2021; Kline et al., 2022; Xiao et al., 2022; Yu et al., 2022).

Early multimodal approaches explored learned feature-to-pixel mappings and dual-CNN architectures for integrating heterogeneous data sources (Sun et al., 2017), followed by hybrid fusion mechanisms that directly embedded tabular information into image representations (Pelka et al., 2020). More recent studies have increasingly relied on feature-level fusion strategies that combine tabular data with latent image representations extracted by deep neural networks (Li et al., 2023), reflecting the ongoing evolution of multimodal AI and heterogeneous data fusion.

Animal bones from archaeological sites are incredibly challenging due to high morphological variability, fragmentation, and the dependence on expert-driven identification methods (Hörr, 2014; Resler et al., 2021). Therefore, the application of AI to zooarchaeology is an underexplored area of research. Nonetheless, the integration of machine learning with geometric morphometrics (GMM), automated landmark detection, and image-based deep learning approaches for the identification of avian skeletal remains (Devine et al., 2020; Klein et al., 2024; Paperini et al., 2026). AI systems may therefore provide valuable support for identifying subtle morphological patterns that are difficult to recognise through traditional approaches alone (Bennett, 2024).

Human–bird relationships play a fundamental role in ecosystem dynamics, subsistence strategies, and cultural practices. The present study was developed within the framework of AviArch, an ERC Consolidator Grant project investigating long-term interactions between humans, birds, and environments through the integration of zooarchaeological analyses, biomolecular approaches, AI-assisted classification of avian remains, and ecological network modelling (https://archaeobiomics.com/aviarch-avifauna-in-archaeoecological-networks/).

In this paper, we present a proof-of-concept multimodal framework for zooarchaeological classification, integrating CNN-based image analysis with standard osteometric measurements. The study develops one model for skeletal element classification and one model for family-level taxonomic identification. Rather than proposing a ready-to-use system for archaeological assemblages, here we establish a controlled methodological baseline based on complete modern reference specimens from selected museum collections. In doing so, it contributes both to the development of a scalable multimodal classification pipeline and to the broader dataset-building effort required for future AI-assisted identification of avian remains under more complex archaeological conditions.

# Dataset and methods

## Dataset and labelling

The data consist of photographs of avian skeletal elements (bird bones) and, when available, osteometric measures. Photographs come from 3 digital collections, having different setups, imaging protocols and measurement devices. The first collection includes about 1800 photographs of 5 skeletal elements from 25 duck species (Haruda et al., 2024) and 5 non-anatid taxa. Most specimens originate from the Zoological Museum of Copenhagen and the Natural History Museum, Tring, and were photographed under controlled conditions. These images are included in (Paperini et al. 2026). The second collection (MPOC - Museo di Geologia e Paleontologia, University of Turin) includes about 7500 images across 12 families, 19 species (selected among those commonly found in Eurasian archaeological assemblages), and 41 individuals, captured under 24 lighting, viewpoint, device and resolution conditions. The third collection, collected by Zooarchaeology specialists at the University of Groningen, adds about 1670 photographs of nine skeletal elements, taken under diverse setups.

In parallel with the image data, 7290 pictures out of 10361 were associated with a vector of standardized osteometric measurements derived from conventional zooarchaeological protocols (von den Driesch, 1976). These measurements included quantitative descriptors of bone morphology, such as linear dimensions of epiphyses, shaft widths, total lengths, and other anatomical metrics that can help limit the potential taxa represented by the bone. The dataset of ranges can be built into the AI classification using data previously published (e.g. Gruwier 2024).

For the classification of bone type, the entire dataset was used; for the family classification, classes with less than 10 images for the training set were not considered, as the small number of images was not sufficient to provide enough examples for the network (see Table 1, Table 2).

Each photograph was renamed to provide information about the specimen, such as binomial nomenclature, genus, collection to which it belongs and indication of bone type: each piece of information was separated by an underscore, following snake case notation.

**Table 1.** Summary of available data for the families.

| Family | Specimens | Bones | Images |
|---|---|---|---|
| Accipitridae | 54 | 130 | 705 |
| Anatidae | 386 | 1306 | 2088 |
| Ciconiidae | 11 | 43 | 464 |
| Columbidae | 6 | 36 | 607 |
| Corvidae | 10 | 46 | 606 |
| Falconidae | 22 | 58 | 523 |
| Gruidae | 11 | 27 | 346 |
| Otididae | 6 | 23 | 111 |
| Phasianidae | 35 | 112 | 887 |
| Rallidae | 24 | 62 | 574 |
| Scolopacidae | 19 | 75 | 682 |
| Strigidae | 31 | 78 | 629 |
| other families / unknown | 1194 | 1608 | 2139 |

**Table 2.** Summary of available data for the bone type.

| Bone type | Specimens | Bones | Images |
|---|---|---|---|
| carpometacarpus | 604 | 605 | 1452 |
| coracoid | 474 | 475 | 1384 |
| cranium | 1106 | 1106 | 1132 |
| femur | 221 | 222 | 1036 |
| humerus | 456 | 457 | 1332 |
| radius | 33 | 34 | 655 |
| scapula | 176 | 177 | 791 |
| tarsometatarsus | 216 | 217 | 1026 |
| tibiotarsus | 226 | 237 | 876 |
| ulna | 35 | 36 | 639 |
| other bones | 38 | 38 | 38 |

# Pre-processing

Prior to segmentation, all images and tabular measurements were standardized using z-score normalization. A measurement-dropout strategy was additionally employed, randomly omitting the osteometric vector during training to improve the robustness of the multimodal framework and enable classification when numerical measurements were unavailable.
Prior to classification, all images were subjected to a two-stage segmentation pipeline. First, BiRefNet (Zheng et al., 2024) was used to identify foreground objects, followed by morphological filtering and a heuristic scoring procedure based on shape, position, and colour to select the most likely bone candidate. The selected region was then analysed in the CIELAB colour space. As an initial quality check, the standard deviation of the B channel (blue–yellow axis) was computed, and regions with a value greater than 5 underwent further refinement, since bone surfaces typically exhibit relatively uniform colour. A two-cluster k-means segmentation was then performed in the LAB colour space, with the a (green–red) and b (blue–yellow) channels weighted twice as much as the L (lightness) channel to prioritize colour information over illumination. Each cluster was assigned a heuristic score that prioritized higher values along the yellow component of the LAB colour space (2x) while giving less weight to overall brightness (0.5x), favouring clusters more likely to correspond to bone, as archaeological clay typically presents grey/blue tones. Refinement was skipped when one cluster was very small, when the segmented region contained fewer than 500 pixels, or when both clusters exhibited elongated shapes, conditions that generally indicated unreliable separation between bone and surrounding material. If these criteria were satisfied, a second segmentation step was performed using SAM2 (Ravi et al., 2024), guided by automatically generated positive and negative prompts. Positive and negative prompt points were automatically selected from the candidate bone and background regions, respectively, and the refinement was applied only when the two prompts were successfully identified and separated by more than 50 pixels. The resulting mask was accepted only if it satisfied predefined plausibility criteria; otherwise, the initial segmentation was retained. Finally, small holes were removed and only the largest connected component was preserved. The segmentation success rate (bone isolated, ruler/background removed) exceeded 95%, the rest of the bones (about 200) were segmented manually.

The selected bone masks were then used to generate the segmented images employed in the subsequent classification pipeline. An overview of the complete preprocessing and multimodal classification workflow is presented in Fig. 1. The resulting masks were used to isolate the bone specimens and generate segmented images for subsequent classification. All code development and execution were carried out in Visual Studio Code using Python 3.12.9 and the PyTorch framework for neural networks.

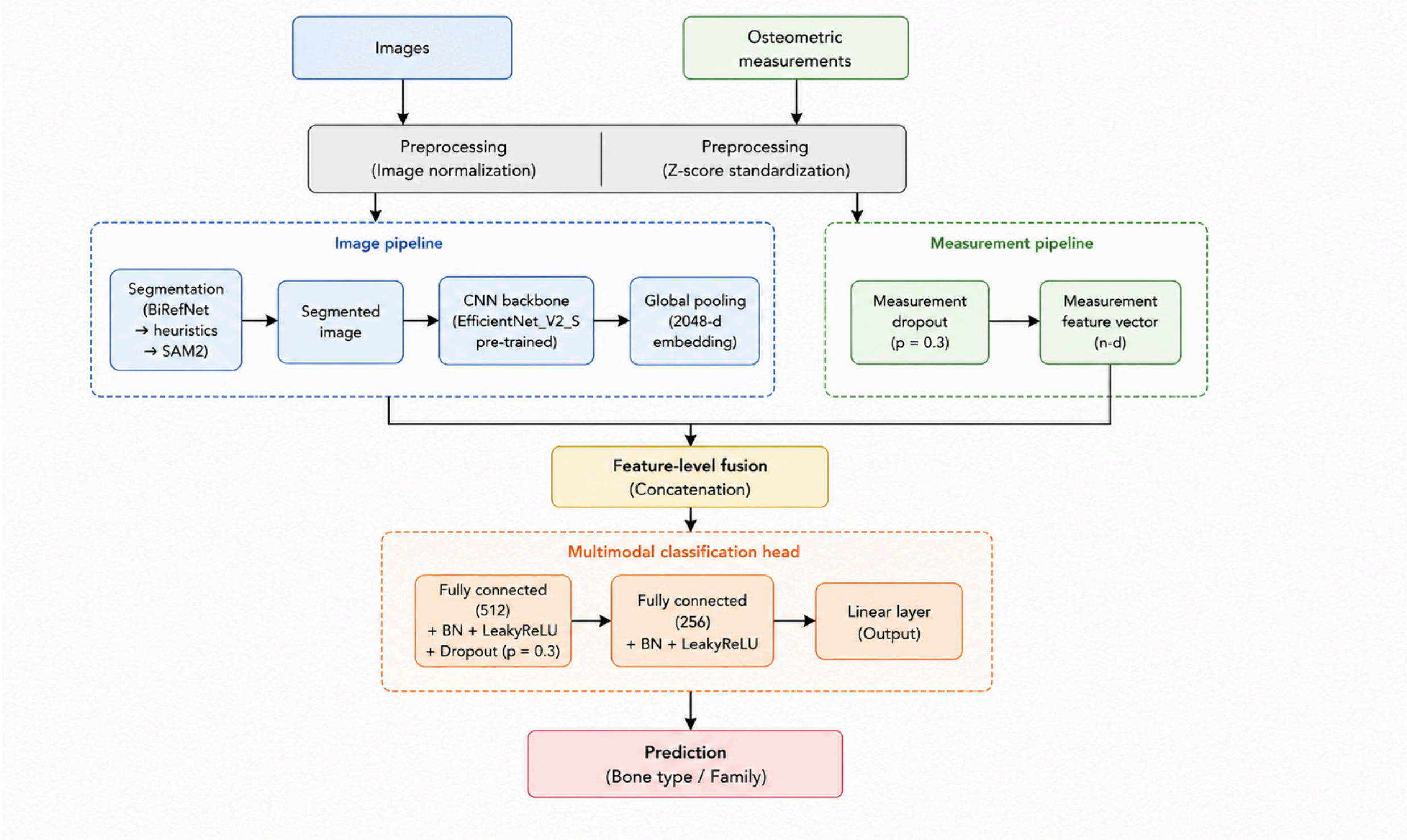

**Fig. 1**. Overview of the proposed multimodal classification framework.

## Classification

Both models were trained and evaluated on a virtual machine equipped with an NVIDIA Tesla T4 GPU, two CPUs, 384 GB of RAM and 800 GB of storage, running on Windows Server 2019. In both cases, transfer learning was employed, using pre-trained networks as backbones.

As illustrated in Fig. 1, the classification framework, for both bone type and family, was implemented using a multimodal architecture combining image-derived deep features with standardized osteometric measurements. The visual branch employed a EfficientNet_V2_S backbone pre-trained on ImageNet (IMAGENET1K_V2 weights).

The original fully connected classification layer of EfficientNet_V2_S was removed and replaced with a custom multimodal classification head. The CNN backbone generated a 2048-dimensional global pooled embedding from each image, while the osteometric variables were processed as a parallel structured feature vector. The two modalities were integrated through feature-level fusion by concatenating the normalized osteometric vector with the image embedding, producing a joint multimodal latent representation encoding both visual morphology and quantitative anatomical information.

The fused representation was subsequently processed through a sequence of fully connected layers designed to progressively reduce dimensionality while preserving discriminative information. Specifically, the concatenated feature vector was projected to 512 units through a fully connected layer followed by batch normalization, LeakyReLU activation, and dropout regularization ($p = 0.3$). A second fully connected block reduced the latent space to 256 dimensions, again followed by batch normalization and LeakyReLU activation. The final linear layer mapped the resulting representation to output logits corresponding to the target families or bone types.

Training was performed in two stages to stabilize optimization and reduce overfitting. During the first stage, the EfficientNet backbone remained frozen and acted exclusively as a fixed feature extractor, while only the multimodal classification head was trained. This strategy allowed the classifier to specialize on osteological and morphometric information without perturbing the generalized visual representations inherited from ImageNet pre-training, thereby limiting overfitting on the relatively small dataset. In the second stage, the final convolutional blocks of the backbone were unfrozen and jointly fine-tuned with the classification head using a reduced learning rate, enabling gradual domain adaptation while preserving previously learned feature hierarchies.

Data augmentation was applied online during training to improve robustness and generalization, using the PyTorch torchvision.transforms.v2 module. Augmentations included random rotations, horizontal and vertical flips, colour jittering, and affine transformations. Model evaluation employed a stratified 15% hold-out test set generated at the bone level, such that all images corresponding to the same bone were assigned to the same partition. The remaining bones were used within a stratified 5-fold cross-validation procedure for model selection and hyperparameter optimization, ensuring robust performance estimation under limited sample conditions.

Optimization was performed using stochastic gradient descent. Differential learning rates were applied to stabilize fine-tuning dynamics: the multimodal classification head was trained using the base learning rate, whereas the unfrozen backbone layers were updated using a learning rate scaled by a factor of 0.1. The classification objective was optimized using categorical cross-entropy loss. Hyperparameter optimization was performed using Optuna with Bayesian optimization to efficiently explore the search space and identify the best-performing configuration.

# Results

Results of the bone type and family classification tasks are presented in Tables 3 and 4. Model performance was primarily evaluated using overall accuracy, defined as the ratio of correctly classified instances to the total number of instances across all classes. For family classification, top-3 accuracy were additionally reported to capture the model's ranking performance and to provide a more practical measure of its predictive robustness. Other measures included precision, recall and F1 score.

**Table 3**. Performance metrics for the model trained to classify bone types. The metrics include Accuracy = (TP + TN) / (TP + TN + FP + FN), Loss, Precision = TP / (TP + FP), Recall = TP / (TP + FN), and F1 Score = 2 * (Precision * Recall) / (Precision + Recall). TP, TN, FP, and FN stand for true positives, true negatives, false positives, and false negatives, respectively.

| Metric | Training set | Validation set | Test set |
|---|---|---|---|
| Accuracy | 0.97 | 0.88 | 0.86 |
| Loss | 0.07 | 0.08 | 0.09 |
| Precision | 0.95 | 0.89 | 0.87 |
| Recall | 0.93 | 0.86 | 0.84 |
| F1 Score | 0.94 | 0.87 | 0.86 |

**Table 4**. Performance metrics for the model trained to classify the family. Metrics formulas are the same as table 3.

| Metric | Training set | Validation set | Test set |
|---|---|---|---|
| Top-1 Accuracy | 0.93 | 0.79 | 0.51 |
| Top-3 Accuracy | 0.98 | 0.87 | 0.75 |
| Loss | 0.25 | 0.75 | 0.98 |
| Precision | 0.92 | 0.78 | 0.53 |
| Recall | 0.91 | 0.77 | 0.50 |
| F1 Score | 0.91 | 0.77 | 0.51 |

Family classification performance was substantially lower than that achieved for bone-type identification. This difference is likely attributable to the greater complexity of the task and the limited amount of training data available for several taxa, which does not fully capture the morphological variability present across species and specimens. The two classification tasks differ considerably in nature: skeletal element recognition relies primarily on anatomical shape, with bone types exhibiting distinct and readily identifiable morphologies. In contrast, family-level identification requires the recognition of more subtle morphological patterns, mirroring the challenges routinely encountered by zooarchaeologists when distinguishing between related taxa.

Nevertheless, when broader ranking metrics are considered, the model achieved a top-3 accuracy of 75%, indicating that the correct family was frequently included among the most probable predictions even when the top-ranked classification was incorrect. This behaviour is particularly relevant in practical zooarchaeological applications, where a restricted set of likely taxonomic candidates can substantially support expert interpretation. Interestingly, despite the greater taxonomic diversity and variability represented in the present dataset, the classification performance remained comparable to that reported by Paperini et al. (2026), whose approach relied exclusively on image-based classification, suggesting that the multimodal framework provides a robust baseline for future developments based on larger and more representative datasets.

# Conclusions

This study presented a proof-of-concept multimodal framework for the automated classification of avian skeletal remains, combining CNN-based visual analysis with osteometric measurements. Rather than proposing a final identification tool, the work establishes a controlled methodological baseline for AI-assisted zooarchaeological classification using complete modern reference specimens. The results show that skeletal element classification can be performed with high reliability, while family-level identification remains more challenging, reflecting the greater subtlety of taxonomic variation and the limited representation of several families in the current dataset.

The contrast between the two tasks is informative. Bone-type recognition relies largely on gross anatomical shape, which is more consistently captured by image-based features. Family classification, by contrast, requires the model to identify finer morphological differences that are often difficult even for specialists, particularly when specimens are closely related. Nevertheless, the relatively strong top-3 performance suggests that the model can provide useful probabilistic support by narrowing the range of plausible identifications, an outcome that is especially relevant for expert-in-the-loop zooarchaeological workflows.

The main contribution of this work is therefore twofold. First, it demonstrates the feasibility of integrating visual and morphometric information within a single multimodal deep learning pipeline for archaeological classification. Second, it contributes to the dataset-building process required to make such systems robust, scalable, and archaeologically meaningful. In this respect, the study aligns with broader developments in multimodal AI for Cultural Heritage, where heterogeneous data sources must be combined to support interpretation rather than treated in isolation.

Several limitations remain. The present models were trained and tested primarily on complete modern reference specimens, and their performance on archaeological remains affected by fragmentation, surface erosion, burning, diagenesis, or missing anatomical features remains to be evaluated. Although all images of the same bone were assigned to a single data partition, the dataset was not split at the specimen level because of the limited number of available individuals, so some degree of shared specimen-specific information between the training and test sets cannot be excluded. In addition, the family-level dataset remained moderately imbalanced despite stratified data partitioning, with several taxa represented by relatively few specimens.

Future work will therefore focus on expanding the dataset to include a broader range of avian taxa and archaeological specimens, improving class balance, and testing the framework under more realistic preservation conditions. Domain adaptation, synthetic damage augmentation, incremental learning, and systematic comparison between image-only, measurement-only, and multimodal models will be important next steps.

More broadly, this study highlights the potential of multimodal AI as a supportive tool for zooarchaeological research, teaching, and reference collection access. Such systems should not replace specialist expertise, but can help structure the identification process, suggest plausible alternatives, and make comparative data more accessible. For this reason, future developments should also address explainability, uncertainty estimation, and data transparency, ensuring that AI-assisted classification remains interpretable, ethically grounded, and useful within archaeological practice.

# Acknowledgments

Reference specimens and image datasets used in this study were provided through collaborations with the Natural History Museum at Tring (United Kingdom), the Natural History Museum of Denmark (University of Copenhagen, Denmark), and the University of Groningen (The Netherlands). We gratefully acknowledge these institutions for granting access to collections, specimens, and associated data that made this research possible.

# Funding

A grant (1024-00032B) from the Independent Research Fund Denmark funded the photography of the Anatidae preliminary dataset. Funding for this research was provided the European Union (ERC-2023-COG HORIZON AviArch, 101125532); the views and opinions expressed, however, are those of the authors only and do not necessarily reflect those of the European Union or the European Research Council. Neither the European Union nor the granting authority can be held responsible for them.